\documentclass[11pt]{article}

\usepackage[preprint]{acl}

\usepackage{times}
\usepackage{latexsym}

\usepackage[T1]{fontenc}

\usepackage[utf8]{inputenc}

\usepackage{microtype}

\usepackage{inconsolata}

\usepackage{graphicx}
\usepackage{amsmath}
\usepackage{mathtools}
\usepackage{amssymb}
\usepackage{amsfonts}
\usepackage{amsthm}

\definecolor{tablecolor}{rgb}{0.8,0.8,0.8}

\newcommand\cut[1]{}

\newcommand{\squishlist}{
   \begin{list}{$\bullet$}
    { \setlength{\itemsep}{0pt}      \setlength{\parsep}{3pt}
      \setlength{\topsep}{3pt}       \setlength{\partopsep}{0pt}
      \setlength{\leftmargin}{1.5em} \setlength{\labelwidth}{1em}
      \setlength{\labelsep}{0.5em} } }

\newcommand{\squishlisttwo}{
   \begin{list}{$\bullet$}
    { \setlength{\itemsep}{0pt}    \setlength{\parsep}{0pt}
      \setlength{\topsep}{0pt}     \setlength{\partopsep}{0pt}
      \setlength{\leftmargin}{2em} \setlength{\labelwidth}{1.5em}
      \setlength{\labelsep}{0.5em} } }

\newcommand{\squishend}{
    \end{list}  }

{}
{}
{}

\newcommand{\myvec}[1]{\mbox{$\mathbf{#1}$}}

\newcommand{\vx}{\mbox{$\myvec{x}$}}

\newcommand{\vy}{\mbox{$\myvec{y}$}}

\newcommand{\calC}{\mbox{${\cal C}$}}

\newcommand{\calH}{\mbox{${\cal H}$}}

\newcommand{\calS}{\mbox{${\cal S}$}}

\newcommand{\vocab}{\mathcal{V}}

\DeclareMathOperator*{\argmax}{arg\,max}

\usepackage{cleveref}
\crefname{section}{Sec.}{Sec.}
\crefname{table}{Tab.}{Tabs.}
\crefname{thm}{Thm.}{Theorem}
\crefname{corr}{Cor.}{Corollary}
\crefname{appendix}{App.}{Appendices}
\crefname{algorithm}{Alg.}{Algorithms}
\crefname{equation}{Eq.}{Eqs.}
\crefname{figure}{Fig.}{Figs.}
\crefname{prop}{Prop.}{Props.}
\usepackage{colortbl}
\usepackage{pgfplots}
\usepackage{subcaption}
\usepackage{tikz}
\usepackage{scalerel}
\usepackage[most]{tcolorbox}
\usepackage{listings}

\definecolor{highlightColor}{HTML}{961C26}
\definecolor{linkColor}{HTML}{00715E}

\definecolor{brandeisblue}{rgb}{0.0, 0.44, 1.0}
\definecolor{brandeisbluecompl}{HTML}{FF8F00}
\definecolor{brandeis2}{HTML}{0F00FF}
\definecolor{brandeis4}{HTML}{00F0FF}
\definecolor{brandeispurple}{HTML}{6D00FF}
\definecolor{palette1}{HTML}{5D85C3}
\definecolor{palette2}{HTML}{005AA9}
\definecolor{palette3}{HTML}{004E8A}
\definecolor{palette4}{HTML}{243572}

\usepackage[most]{tcolorbox}
\usepackage{listings}
\tcbuselibrary{breakable}
\newtcolorbox{promptbox}{
  breakable,
  colback=gray!5,
  colframe=gray!60,
  boxrule=0.5pt,
  arc=2pt,
  left=6pt,
  right=6pt,
  top=6pt,
  bottom=6pt,
}

\title{Uncertainty-Aware Generation and Decision-Making Under Ambiguity}

\author{Nico Daheim \quad \quad 
  Iryna Gurevych \\
  Ubiquitous Knowledge Processing Lab (UKP Lab),\\ Department of Computer Science, Technical University of Darmstadt \\ National Research Center for Applied Cybersecurity ATHENE, Germany\\
  \href{http://www.ukp.tu-darmstadt.de/}{www.ukp.tu-darmstadt.de} \\}

\begin{document}
\maketitle

\begin{abstract}
{\color{red} Disclaimer: We do not promote automation of the peer-review process but aim to support human reviewers and authors.}

With rapidly improving capabilities, Large Language Models (LLMs) are increasingly used in many complex real-world tasks.
Beyond requiring in-depth knowledge and reasoning skills, many of these tasks exhibit a high degree of subjectivity and require that the outputs of the model can be trusted.
While a lot of progress has been made to train better models, decision-making algorithms have received less attention.
In this work, we present and evaluate various uncertainty-aware decision-making algorithms based on Bayesian decision theory and risk-averse decision making on the tasks of tutoring and automatic peer reviewing.
Concretely, we take uncertainty over tutoring strategies and review scores into account when generating a tutor response or review and use conformal prediction to provide guarantees over strategy and score.
We find empirically that these algorithms can improve the utility of the generations but need to be carefully implemented when ambiguity is high.
For example, risk-averse rules can degrade performance by optimizing for generic outputs, while Bayesian methods tend to perform better.
Our work uses techniques from decision theory to improve LLM-based decision-making and outlines open challenges for the community.\footnote{Code available under \url{https://github.com/UKPLab/arXiv2026-uncertainty-aware}.}
\end{abstract}

\section{Introduction}
\begin{figure}[t]
    \includegraphics[width=\linewidth,trim=7in 3in 7in 2.75in,clip]{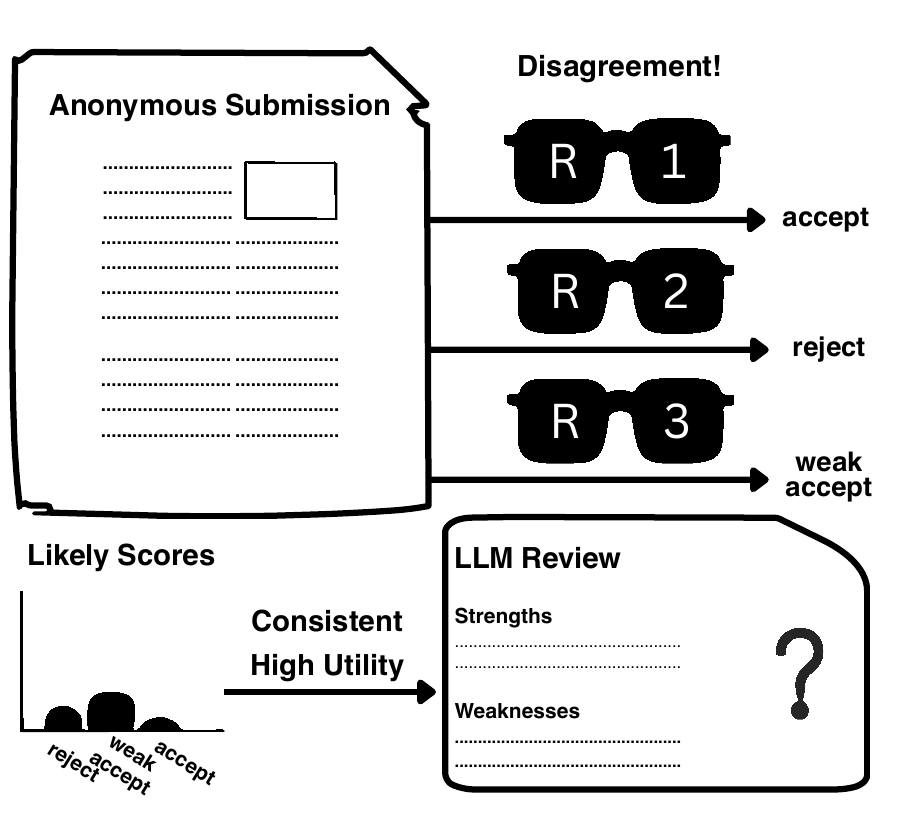}
    \caption{In many real-world tasks there is a large amount of uncertainty over decisions and even humans disagree significantly.
             For example, in reviewing human reviewers often disagree about a paper's scores.
             We argue that automatic methods for supporting humans in these tasks should take this uncertainty into account to generate high-utility outputs while narrowing down plausible ranges of intermediate decisions.}
\end{figure}
By now, LLMs are ubiquitously used, including in applications that were deemed out of reach just years ago because they are sensitive and require expert-level knowledge.
One example is tutoring, where LLMs have now already found their way into real-world applications~\citep{learnlmteam2025learnlmimprovinggeminilearning}.
Another example is peer review, where several sub-communities in machine learning research are adopting LLM-based tools to support authors and reviewers.
NeurIPS 2026 uses the tool of~\citet{cohen-addad2025gemini} to provide LLM-generated reviews to authors before submission~\citep{neurips_pat_2026}.
The tool is based on Gemini-2.5~\citep{comanici2025gemini} and has received overwhelmingly positive feedback in a pilot at the STOC 2026 conference.
ACL Rolling Review already uses the Revas tool to support reviewers.\footnote{\url{https://revas.mbzuai.ac.ae/}}
Overall, LLMs are increasingly used in many such high-stakes domains.

These domains bring up a list of challenges that need to be taken care of.
For example, if students are told solutions too early their learning can be slower than with active learning approaches that allow exploration by the student~\citep{Chi02102014}. %
In peer review, incorrect or imprecise feedback might lead authors to add revisions that are against conference guidelines, and misleading scores could lead to authors submitting early versions of works that are very unlikely to be accepted and overload peer review even more.

However, addressing these challenges is often hard for both humans and models.
One reason for that is that both these settings involve multiple decision steps.
A teacher needs to find out where the student made a mistake, and then pick a strategy to address it.
A reviewer needs to assess many nuanced points about a paper and then provide both a review and a score.
In addition, both settings are highly ambiguous and even different human teachers or reviewers will pick fundamentally different (but valid) strategies or scores and might even contradict each other~\citep{kumar-etal-2023-reviewers}.

A variety of works have been proposed that change the model, for example, to split tasks into multiple sequential reasoning steps before a final answer generation.
In tutor response generation this usually entails finding a student mistake, picking a strategy and responding~\citep{wang-etal-2024-bridging, daheim-etal-2024-stepwise,hsu-etal-2026-mathedu}.
In reviewing, multiple steps might be identifying novelty~\citep{afzal-etal-2026-beyond} and finding suggestions to present~\citep{zhu-etal-2025-deepreview}.
The same is true in related domains like NLP for mental health, where intermediate reasoning can help in applications like reframing~\citep{goel-etal-2025-socratic}.
Other works try to improve models via reinforcement~\citep{dinucu-jianu-etal-2025-problem} or supervised learning~\citep{idahl-ahmadi-2025-openreviewer} or use multi-agent systems~\citep{weng2026deepreviewer20traceableagentic}.
Many of these problems can be formulated as the prediction of a state and response but common decision-making algorithms usually neither account for ambiguity over states nor can provide any guarantees over them. This often limits their real-world utility, for example, because they can not narrow down plausible score ranges well.

Here, we present decision-making and generation algorithms that can both account for uncertainty and provide statistical guarantees and highlight various open challenges for the community.
First, we use a Bayesian approach that maximizes an expected utility for an LLM generation with respect to a distribution over such a state.
In tutoring, this state is the tutoring strategy and in reviewing this is the reviewer score.
We use conformal prediction to provide a statistical guarantee over this state, which is also used to prune the number of states that are considered for utility estimation.
Finally, we use a risk-averse strategy~\citep{kiyani2025decision} that maximizes the utility of the worst-case state in this conformal prediction set.
These strategies can be used to improve the predictions of any model that predicts both state and response.

We evaluate these methods on the tasks of review generation and scoring as well as on the task of tutor response generation in connection with tutoring strategies.
For generation, we use Gemma-3-27B~\citep{gemmateam2025gemma3technicalreport}, Qwen3-30B~\citep{yang2025qwen3technicalreport} and Mistral-3.2-24B~\citep{mistralai2025mistral3} for reviewing, and TutorRL-7B for tutoring~\citep{dinucu-jianu-etal-2025-problem}.
Across multiple datasets, namely, Review-5K~\citep{weng2025cycleresearcher}, NLPEER~\citep{dycke-etal-2023-nlpeer}, and MathDial~\citep{macina-etal-2023-mathdial} we find that these methods can increase the utility of reviews and tutor responses.
We also find that Bayesian strategies seem to be more robust to the quality of state distribution and that meaningful prediction sets depend on both classifier performance and how ambiguity is handled.
Overall, our work outlines methods for uncertainty-aware decision-making with LLMs as well as open problems for the community.

\section{Background}

\subsection{Tutor Response Generation}
The goal of tutor response generation is to generate a response $\vy = (y_1, \dots, y_T) \in \vocab^\ast$ from a vocabulary $\vocab$ given a set of inputs that we will refer to as $\vx \in \vocab^\ast$ for generality.
Usually, these inputs include the previous dialog history, supporting knowledge, an incorrect solution step, and a given problem.
The amount of information that the LLM has to process in order to solve the task already hints that the task is quite hard to solve directly.
Just prompting an LLM with $\vx$ as the input and sampling $\vy \sim p(\cdot\mid \vx)$ often leads to the LLM providing incorrect feedback~\citep{macina2023opportunities} or telling the solution to the student early on, which can be detrimental to their learning.
Other strategies should be preferred that provide the student with room to explore possible problem solutions in order to foster learning~\citep{lepper2002wisdom, nye2014autotutor}, %
sometimes going as far as letting the student fail proactively in order to increase exploration~\citep{kapur2008productive}.

To reduce the amount of incorrect feedback and improper tutoring strategies, many recent works model tutoring as a multi-step problem.
For example,~\citet{wang-etal-2024-bridging} classify the type of error a student has made given a pre-defined taxonomy of errors and then predict a tutoring strategy and intent.
~\citet{daheim-etal-2024-stepwise} propose to first verify the student solution attempt with free-text reasoning created by an LLM that is then used to produce a response.
Another example is the work by~\citet{puech-etal-2025-towards} who trace student solutions and then classify an intent before generating a response.
Such strategies have been adopted in a variety of works~\citep{hsu-etal-2026-mathedu, park-etal-2026-beyond} but, notably, are prone to propagating errors in initial steps.

Other works introduce training and data collection strategies to improve tutoring models, for example, using reinforcement learning~\citep{dinucu-jianu-etal-2025-problem} or by collecting synthetic data~\citep{learnlmteam2025learnlmimprovinggeminilearning}.
Finally, methods have been proposed to evaluate how well the responses generated by a tutor follow pedagogical principles~\citep{macina-etal-2025-mathtutorbench}.

\subsection{Automated Methods in Peer Review}
Similar observations have been made in the area of automated peer review.
There, the goal is to generate a review $\vy \in \vocab^\ast$ given a context $\vx$ that usually consists of a research paper and, perhaps accompanying materials.
Oftentimes, a review score is also predicted which can be useful, for example, for authors to understand their chances of acceptance.
Our aim is to explore such review generation for peer review support, for example, to provide meaningful critiques to authors before submission or help reviewers write better reviews.

Due to its complexity, recent works also split reviewing into multiple steps.
For example,~\citet{weng2025cycleresearcher} generate multiple reviews sequentially using the same LLM that are then summarized to mimic a senior reviewer.
Another example is the work by~\citet{zhu-etal-2025-deepreview}, who first identify novelty, then generate and verify reviews, and then generate a meta-review and score.
Recently, such multi-step systems have also been modeled with multiple agents that might take on specific tasks and roles~\citep{gao2025reviewagentsbridginggaphuman, weng2026deepreviewer20traceableagentic}.

Other works look at some of these steps in isolation.
For example,~\citet{afzal-etal-2026-beyond} try to identify novelty in scientific papers based on a multi-step procedure and~\citet{purkayastha-etal-2026-decision} focus on modeling meta reviewing.
Recently, there have also been works to evaluate scientific reviews according to a set of pre-defined criteria~\citep{sadallah-etal-2025-good, sahinuc2026reward}.

At the same time, decision-making and generation algorithms have received less attention in both domains.
As a consequence, while there are models and approaches to understand the utility of a review, it is not maximized.
Furthermore, uncertainty over decisions, for example, over the right score or tutoring strategy is not modeled explicitly.
Errors may be propagated and there are no indications of error rates which can endanger user trust~\citep{dhuliawala-etal-2023-diachronic, palod2026evaluatingfalsetrustengendered}.
Ideally, we would like a decision-making algorithm that both accounts for uncertainty and provides statistical guarantees over states to increase trust.

\subsection{Uncertainty-Aware Methods in NLP}
Many works have been proposed to account for utility and uncertainty, and to provide guarantees, but these are often believed to be conflicting.
As a consequence, methods usually provide one or the other, but not both at the same time.

Minimum Bayes Risk decoding~\citep{kumar-byrne-2004-minimum} is one method to account for utility in the sense of a comparison between one output $\vy$ and another output $\vy^\prime$ via a utility function $u(\vy, \vy^\prime)$ and the following decision rule, \begin{equation}
    \vy^\ast = \argmax_{\text{\vy} \in \vocab^\ast} \sum_{\text{\vy}^\prime\in\vocab^\ast}p(\vy^\prime\mid \vx) \cdot u(\vy, \vy^\prime).
\end{equation}
The objective is usually approximated by a collection of samples, which we will denote with $\calH$, for example, using Monte-Carlo methods~\citep{eikema-aziz-2022-sampling} or the model probability directly~\citep{jinnai2024modelbased}, and can be extended to ensembles~\citep{daheim2025uncertaintyaware}.
However, the choice of $u$ is not easy and the method scales quadratically with the size of $\calH$, though methods to reduce this exist~\citep{eikema-aziz-2022-sampling, cheng-vlachos-2023-faster}.
To address these, we here adopt a similar strategy but target a different decision problem that does not compare different $\vy$.
The method is similar to quality-aware decoding~\citep{fernandes-etal-2022-quality} and best-of-n sampling~\citep{stiennon2020learning} and a similar formulation has been explored by~\citep{johnson2023rusure} for code generation.

Split conformal prediction~\citep{vovk2005algorithmic} provides what is referred to as marginal coverage for a classifier, which means that \begin{equation}
    P(\widehat{\vy} \in \calC(\vx)) \geq 1 - \alpha,
\end{equation}
given a threshold $\alpha$, ground-truth label $\widehat{\vy}$ and prediction set $\calC(\vx)$ for a random (not fixed) $\vx$ in a test set.
In words, for exchangeable data (see~\citet[Sec. 3.1]{campos-etal-2024-conformal} for a precise definition), with probability at least $1-\alpha$ the correct label is contained in $\calC(\vx)$.
The prediction set is constructed using a threshold determined on a separate validation set with the same classifier.
The formulation has been adopted and built upon in many general machine learning and NLP works~\citep{angelopoulos2023gentle, campos-etal-2024-conformal}.

Historically, there is a lot of discussion about whether Bayesian methods and frequentist methods like conformal prediction are reconcilable, see e.g.~\citet{robert2007bayesian}.
Here, we do not intend to add to this discussion but intend to show that combining such techniques can achieve many real-world goals when applying LLMs to complex tasks.

\section{Bayes Decision Rule}
\label{sec:bayes_decision_rule}
We discuss decision problems with two components.
The first component is a decision space $\vocab^\ast$ that coincides with the space of all strings that can be generated using a vocabulary $\vocab$ and contains reviews for academic works and responses of tutors.
The second component is a finite, discrete, and problem-specific set $\Omega$. For example, in tutoring this is a set of tutoring strategies and in reviewing this could be possible scores according to a scoring guideline of a conference. That is, in reviewing each $\omega \in \Omega$ is a specific score.
The decision is taken given $\vx \in \vocab^\ast$ which is contextual information like the paper in peer review or dialog context in tutoring.
In both cases, we aim to find a final output $\vy^\ast \in \vocab^\ast$ that should have high utility and account for uncertainty over choices of $\omega \in \Omega$ which often arises due to ambiguity.
How utility is defined is problem-specific and, ideally, we would like to empower practitioners to be able to define and use their own criteria.

We argue that Bayesian decision theory provides a rule that satisfies these desiderata for $\vy^\ast$.
The rule is to choose the output $\vy^\ast$ that minimizes the Bayes risk or, equivalently, maximizes expected utility~\citep[Sec. 8.2]{degroot2005optimal}, as follows,\begin{equation}
    \vy^\ast = \argmax_{\text{\vy} \in \vocab^\ast} \sum_{\omega \in \Omega} u(\vy, \omega) \cdot p(\omega\mid \vx).
    \label{eq:bayes_risk}
\end{equation}
The corresponding $\vy^\ast$ is then called Bayes decision against the distribution $p(\omega\mid \vx)$.
The rule (in expectation) guarantees optimality wrt. $u$ and $p(\omega\mid \vx)$ but does not guarantee optimal decisions for another utility function.
It is important to choose it wisely to avoid metric bias~\citep{kovacs-etal-2024-mitigating}.

Dissecting~\cref{eq:bayes_risk} shows that $u(\vy, \omega)$ is a function that depends on both a specific choice $\omega \in \Omega$ and a specific output $\vy\in \vocab^\ast$ which can be used to encode both quality and consistency criteria.
Concretely, in tutoring we can give a score that indicates for each tutor response $\vy$ whether it follows sound pedagogical principles and whether it is consistent with a specific tutoring strategy $\omega$.
Responses that are consistent with likely tutoring strategies are preferred, and so are responses that follow sound pedagogical principles.
Crucially, we do not make a hard decision according to a specific tutoring strategy but actively reduce uncertainty over them via marginalization.

In~\cref{eq:bayes_risk}, there is no notion of output sequence probabilities but these can be incorporated by using Bayes theorem.
For example, we may rewrite \begin{equation}
    \begin{split}
        p(\omega\mid \vx) &\propto \sum_{\text{\vy}\in\vocab^\ast} p(\omega\mid \vy, \vx)\cdot p(\vy\mid \vx)\\ 
    \end{split}
    \label{eq:bayestheorem}
\end{equation}
to include $p(\vy\mid \vx)$.
Many choices can be made by the practitioner that can have different advantages in practice.
These choices are important, because $\cref{eq:bayes_risk}$ can not be calculated exactly due to expectations and/or maximization over the countably-infinite space $\vocab^\ast$ and corresponding estimators do not need to yield the same decisions, even if they estimate equivalent rewritings.
One option is to evaluate $p(\omega\mid \vx)$ directly and sample $\calH = (\vy^{(1)}, \dots, \vy^{(M)} \mid \vy^{(i)} \sim p(\cdot \mid \vx))$ to restrict the maximization to a promising subset, 
\begin{equation}
    \widehat{\vy}^\ast = \argmax_{\text{\vy} \in \text{\calH}} \sum_{\omega \in \Omega} u(\vy, \omega) \cdot p(\omega\mid \vx).
    \label{eq:bayes_risk_estimator}
\end{equation}
The formulation is similar to best-of-n sampling~\citep{stiennon2020learning} but reduces uncertainty over an additional state space $\Omega$ via smoothing according to $p(\omega\mid \vx)$.
Another option is to use conditional distributions by either conditioning the LLM on a categorical output or conditioning the model of $\Omega$ on outputs.
For example, we could generate responses for specific tutoring strategies.
These could use a Monte-Carlo approximation of the sum in~\cref{eq:bayestheorem} but it is also possible to use the LLM probabilities directly~\citep{jinnai2024modelbased}.

The method is clearly related to Minimum Bayes Risk decoding~\citep{goel2003minimum} but both aim to minimize the Bayes risk of fundamentally different decision problems.
Usually, in MBR the space $\Omega$ is chosen as $\Omega = \vocab^\ast$ and as a consequence the utility function compares two different output strings.
While this is valid and also successful in practice for domains like machine translation~\citep{freitag-etal-2022-high}, there are also issues.
It is often hard to define $u(\vy, \vy^\prime)$ and, indeed, multiple works instead use pointwise utility functions that do not account for uncertainty via marginalization anymore~\citep{fernandes-etal-2022-quality}.
These can also be used to prune collections $\calH$ as MBR has, in general, a quadratic complexity in $|\calH|$.
~\cref{eq:bayes_risk} has much lower complexity if $|\Omega| << \calH$.
Our method is also inspired by~\citet{johnson2023rusure}, who use a decision rule like~\cref{eq:bayes_risk} to find good code suggestions.
We build upon this work and show further connections to other uncertainty-aware methods and expand the discussion to general problems.

\section{Conformal Prediction \& Risk Aversion}
The rule in~\cref{eq:bayes_risk} tells us how to choose $\vy^\ast$ but is perhaps less informative about $\Omega$.
The Bayes decision for selecting $\omega^\ast$ wrt. classification accuracy is to pick the maximizer of $p(\omega \mid \vx)$ but presenting only the corresponding $\omega^\ast$ does not provide a practitioner with an uncertainty estimate over the entire space and the distribution $p(\omega \mid \vx)$ might or might not be informative. For example, we found that Qwen3-14B assigned probabilities close to 1 for one class and close to 0 for other classes when prompted zero-shot in review score prediction.
Ideally, an author submitting their paper to an automatic review before conference submission would like to know plausible intervals of scores to understand the acceptance chances of their paper.
Such a method would also help the community, as papers that are likely to receive lower scores might be less likely to be submitted which could reduce the overall number of submissions.
For tutoring, it would help, for example, to understand common patterns in tutoring strategies over a full dialog.

Conformal Prediction provides such a method which produces prediction sets $\calC(\vx)$ that, measured over a random test set, cover the true labels with probability at least $1 - \alpha$.
In split conformal prediction~\citep{vovk2005algorithmic}, this is achieved by splitting the data into a training set to obtain a classifier over $\Omega$ and a validation set that is used to determine a threshold $\tau$ based on a non-conformity score, for which we use $1-p(\omega\mid \vx)$.
Still, there are some issues with split conformal prediction in our setting, because it assumes that data is exchangeable and that one ground-truth label exists.
These assumptions are violated, for example, when data distributions drift and data is not identically distributed.
Importantly, the assumptions also may not hold when the label is ambiguous~\citep{stutz2023conformal} under human label variation~\citep{plank-2022-problem}.

Arguably, this is the case in all of the applications we discuss: different reviewers score the same paper differently and different teachers prefer different but similarly effective tutoring strategies.
In such a case, standard conformal prediction may not provide the same $1-\alpha$ coverage.
Instead,~\citet{stutz2023conformal} propose Monte-Carlo conformal prediction which uses an augmented set for calibration, where the score $1-p(\omega\mid \vx)$ is calculated multiple times for same input $\vx$ but for different $\omega$. 
These can either be based on an existing set of labelings or on samples from an estimated label distribution with a slight modification of the quantile definition.
Then, coverage can only be guaranteed wrt. this distribution and theoretical coverages are widened to $1-2\alpha$, though empirically a tighter $1-\alpha$ is often observed.
We choose this strategy whenever multiple labels are available and otherwise fall back to using one label.
We emphasize this as an important problem for the community.
Instead of only a single label, multiple labels should be collected to be able to explicitly consider label variation.

Here, we use conformal prediction to find $\calC(\vx) \subseteq \Omega$ that can be used to reduce the number of utility calculations in~\cref{eq:bayes_risk} if $|\calC(\vx)| < \Omega$ and provide a set of plausible scores or tutoring strategies that can be used for further insights.
This is similar to~\citet{cheng-vlachos-2023-faster} but uses calibrated scores.
Applied to~\cref{eq:bayes_risk} this means, \begin{equation}
\vy^\ast = \argmax_{\text{\vy} \in \vocab^\ast} \sum_{\omega \in \calC(\vx)} u(\vy, \omega) \cdot p(\omega\mid \vx).
\label{eq:bayesrisk_conformal}
\end{equation}
The objective also bridges to the risk-averse decision rule from~\citet[Eq. 8]{kiyani2025decision} if $p(\omega\mid \vx)$ is chosen to be a Dirac distribution that places all mass on the worst-case $\omega \in \calC(\vx)$ in the sense of the utility $u$.
The corresponding minmax decision is given by the decision rule\begin{equation}
\vy^\ast = \argmax_{\text{\vy} \in \vocab^\ast} \min_{\omega \in \calC(\vx)} u(\vy, \omega)
\label{eq:minmax}
\end{equation}
and can be seen as maximizing a worst-case utility.
For example, in reviewing this maximizes the utility of a worst-case score to reduce the chance of presenting a poor-quality review to the user.
Again, the same estimators as discussed in~\cref{sec:bayes_decision_rule} can be used to narrow down promising examples for the maximization over $\vocab^\ast$.
\begin{table*}[t]
    \centering
    \resizebox{\linewidth}{!}{
        \begin{tabular}{l|cccc|cccc}
                & \multicolumn{4}{c}{ICLR-5k} & \multicolumn{4}{c}{NLPEER} \\
         Method & Actionable & Verifiable & Helpful & Grounded  & Actionable & Verifiable & Helpful & Grounded  \\ \hline
                &  \multicolumn{8}{c}{Gemma3-27B} \\ \hline
        Sampling & 1.99\tiny $\pm$0.008 & 2.44\tiny $\pm$0.007 & 2.80\tiny $\pm$0.002 & 3.02\tiny $\pm$0.005 & 1.89\tiny $\pm$0.018 & 2.36\tiny $\pm$0.013 & 2.74\tiny $\pm$0.004 & 2.92\tiny $\pm$0.011 \\
        \rowcolor{gray!10} Bayes & 2.13\tiny $\pm$0.009 & 2.48\tiny $\pm$0.005 & \bf 2.84\tiny $\pm$0.003 & 3.06\tiny $\pm$0.004 & \bf 2.15\tiny $\pm$0.020 & \bf 2.45\tiny $\pm$0.013 & \bf 2.82\tiny $\pm$0.012 & \bf 3.05\tiny $\pm$0.011 \\
        \rowcolor{gray!10} + Conformal & \bf 2.14\tiny $\pm$0.005 & 2.48\tiny $\pm$0.007 & \bf 2.84\tiny $\pm$0.002 & \bf 3.07\tiny $\pm$0.005 &  2.13\tiny $\pm$0.018 & \bf 2.45\tiny $\pm$0.011 & \bf 2.82\tiny $\pm$0.008 & 3.04\tiny $\pm$0.010\\
        \rowcolor{gray!10} Minmax & 2.04\tiny $\pm$0.014 & \bf 2.49\tiny $\pm$0.010 & 2.83\tiny $\pm$0.013 & \bf 3.07\tiny $\pm$0.011 & 1.93\tiny $\pm$0.017 & 2.39\tiny $\pm$0.012 & 2.77\tiny $\pm$0.008 & 3.00\tiny $\pm$0.019 \\ \hline
        & \multicolumn{8}{c}{Mistral3-24B} \\ \hline
        Sampling & 1.92\tiny $\pm$0.017 & 2.37\tiny $\pm$0.012 & 2.77\tiny $\pm$0.003 & 2.68\tiny $\pm$0.007 & 1.79\tiny $\pm$0.030 & 2.30\tiny $\pm$0.016 & 2.70\tiny $\pm$0.014 & 2.63\tiny $\pm$0.010 \\
        \rowcolor{gray!10} Bayes & \bf 2.26\tiny $\pm$0.021 & \bf 2.53\tiny $\pm$0.010 & 2.86\tiny $\pm$0.015 & \bf 2.77\tiny $\pm$0.007 & \bf 2.18\tiny $\pm$0.015 &\bf 2.53\tiny $\pm$0.012 & \bf 2.85\tiny $\pm$0.008 & \bf 2.77\tiny $\pm$0.016\\
        \rowcolor{gray!10} + Conformal & \bf 2.26\tiny $\pm$0.019 & \bf 2.53\tiny $\pm$0.005 & \bf 2.87\tiny $\pm$0.014 & \bf 2.77\tiny $\pm$0.006 & 2.14\tiny $\pm$0.019 & 2.51\tiny $\pm$0.009 & 2.84\tiny $\pm$0.010 & 2.76\tiny $\pm$0.015\\
        \rowcolor{gray!10} Minmax & 2.13\tiny $\pm$0.024 & 2.44\tiny $\pm$0.015 & 2.82\tiny $\pm$0.013 & \bf 2.77\tiny $\pm$0.009 & 1.93\tiny $\pm$0.035 & 2.30\tiny $\pm$0.016 & 2.69\tiny $\pm$0.022 & 2.65\tiny $\pm$0.009 \\ \hline
        & \multicolumn{8}{c}{Qwen3-30B} \\ \hline
        Sampling & 1.59\tiny $\pm$0.030 & 2.84\tiny $\pm$0.009 & 2.77\tiny $\pm$0.011 & 3.28\tiny $\pm$0.018 & 1.48\tiny $\pm$0.009 & 2.77\tiny $\pm$0.038 & 2.77\tiny $\pm$0.033 & 3.09\tiny $\pm$0.020 \\
        \rowcolor{gray!10} Bayes & \bf 1.69\tiny $\pm$0.014 & 2.88\tiny $\pm$0.011 & 2.80\tiny $\pm$0.005 & 3.32\tiny $\pm$0.016 & \bf 1.55\tiny $\pm$0.025 & 2.84\tiny $\pm$0.049 & \bf 2.86\tiny $\pm$0.037 & \bf 3.16\tiny $\pm$0.020 \\
        \rowcolor{gray!10} + Conformal & 1.68\tiny $\pm$0.015 & 2.87\tiny $\pm$0.011 & 2.79\tiny $\pm$0.003 & 3.31\tiny $\pm$0.015 & 1.53\tiny $\pm$0.016 & 2.84\tiny $\pm$0.060 & \bf 2.86\tiny $\pm$0.046 & 3.15\tiny $\pm$0.036\\
        \rowcolor{gray!10} Minmax & 1.67\tiny $\pm$0.006 & \bf 2.96\tiny $\pm$0.004 & \bf 2.82\tiny $\pm$0.002 & \bf 3.36\tiny $\pm$0.015 & 1.35\tiny $\pm$0.029 & \bf 2.89\tiny $\pm$0.010 & 2.84\tiny $\pm$0.009 & 3.05\tiny $\pm$0.006\\ \hline
    \end{tabular}
    }
    \caption{We compare~\cref{eq:bayes_risk} (Bayes) with and without a pruning via conformal prediction to~\cref{eq:minmax} (Minmax) on ICLR and ARR reviews from NLPEER. 
    We use the reward models from~\citet{sahinuc2026reward} that assign a score from 1-4 to each aspect.
    Both methods provide improvements, with Bayes usually performing better, except for the case of Qwen3, where Minmax provides stronger improvements in some criteria.
    Averaged over 3 seeds.
    }
    \label{tab:tab1}
\end{table*}

\section{Experiments}
We conduct experiments on review generation and scoring as well as on tutor response generation and tutoring strategy prediction.
Both times, we compare sampling against the Minimum Bayes Risk objective from~\cref{eq:bayes_risk} with and without a conformal set pruning and the minmax objective from~\cref{eq:minmax}.

We use MathDial~\citep{macina-etal-2023-mathdial} for tutoring and NLPEER~\citep{dycke-etal-2023-nlpeer} and ICLR submissions from Review-5k~\citep{weng2025cycleresearcher} for reviewing.
We use \textrm{google/gemma-3-27b-it}~\citep{gemmateam2025gemma3technicalreport}, \textrm{Qwen/Qwen3-30B-A3B-Instruct-2507}~\citep{yang2025qwen3technicalreport}, and \textrm{mistralai/Mistral-Small-3.2-24B-Instruct-2506}~\citep{mistralai2025mistral3} to generate reviews conditioned on the paper and the scoring guidelines from ICLR and ACL Rolling Review.
For tutoring, we use \textrm{TutorRL-7B}~\citep{dinucu-jianu-etal-2025-problem} for generation and prompt it with the previous dialog history, problem, student solution attempt, and ground-truth solution.

Due to finding poor zero-shot prompting performance, we train \textrm{meta-llama/Llama-3.2-3B-Instruct}~\citep{grattafiori2024llama} to predict a distribution over $\Omega$. 
We use $\alpha=0.05$ for conformal prediction.
For $u$, we use Qwen3-14B as an LLM judge with 5-point Likert-scale ratings defined in~\cref{sec:details}.
We use an expectation over rubric scores as concurrently proposed by~\citet{kwok2026llmverifier}.
We use the reward model from~\citet{macina-etal-2025-mathtutorbench} to evaluate tutor responses and the reward model from~\citet{sahinuc2026reward} to evaluate reviews.
Both are different from the model used to calculate utility and have been shown to outperform prompted LLMs.

All experimental details are in~\cref{sec:details}, where we also report the exact prompts that are used for generation, scoring of categorical labels, and utility estimation.
Unless otherwise stated, we use $|\calH| = 64$ for reviewing and $|\calH| = 32$ for tutoring. 
 
\section{Results}

\subsection{Main Results on Reviewing}
We first show our main results on the task of review generation and scoring in~\cref{tab:tab1} on Review-5k and NLPEER, where we use the subset of papers with more than one review for testing.
We compare the Bayes decision rule from~\cref{eq:bayes_risk} to using a conformal pruning as in~\cref{eq:bayesrisk_conformal} and using the minmax rule from~\cref{eq:minmax}.
$\calH$ is sampled from the LLM with ancestral sampling and temperature $1$ and we disable all truncations of token-level distributions.

We use a finetuning of Llama-3.2-3B-Instruct on the training set of Review-5k and papers with one review in NLPEER, as we found that zero-shot prompting of larger models like Qwen3-14B was insufficient for score prediction and, for example, only obtained an accuracy of ca. 24\% on NLPEER.
In both cases we split $90\%$ for training and $10\%$ for calibration.
For reviewing, the score distribution $p(\omega\mid \vx)$ is estimated by predicting the scores of each generated review and averaging all resulting score distributions for the same paper.
The resulting classifiers obtained $32.6\%$ accuracy on the Review-5k validation set and $30.1\%$ on the NLPEER validation set which highlights the large amount of ambiguity and might also be due to the shift from human reviews to generated reviews during prediction.
We have also tried using an oracle based on the human reviews which achieved better results ($58.6\%$ on Review-5k) but downstream generation metrics did not differ significantly.

In our results, we find that both methods generally give improvements over just sampling responses.
Overall, the improvements with the Bayes decision rule seem to be the largest, especially in terms of actionability.
We believe that this is due to the prompt used for utility estimation that names actionability first and which might thus be favored.
Conformal pruning seems to mostly retain performance at the benefit of requiring fewer utility estimations.
The savings are shown in~\cref{tab:tab3}.
The minmax rule performs similarly. Actionability is often lower but, interestingly, on Qwen3 in terms of other criteria it performs best on Review-5k.

When comparing models, we find that Gemma3 and Mistral3 perform similarly.
Qwen3 is not as good in actionability but better in other criteria which might be due to the post-training of Qwen3 favoring these criteria.
Our results show many open avenues for future work.
First, future work can look into better LLMs for sampling and better utility estimation.
In addition, future work should look into better score prediction methods that might further improve performance.
\begin{table}[t]
    \centering
    \resizebox{\linewidth}{!}{\begin{tabular}{l|ccc}
         Method & Reward & Win (Sampl.) & Win (GT) \\ \hline
         Sampling & 0.858 \tiny $\pm$ 0.002 & - & 71.6 \tiny $\pm$ 0.01\\
         \rowcolor{gray!10} Bayes & \bf 0.892 \tiny $\pm$ 0.002 & \bf 54.3 \tiny $\pm$ 0.09 & \bf 75.4 \tiny $\pm$ 0.04 \\
         \rowcolor{gray!10}+ Conformal  & \bf 0.892 \tiny $\pm$ 0.002 & \bf 54.3 \tiny $\pm$ 0.01 & \bf 75.2 \tiny $\pm$ 0.04\\
         \rowcolor{gray!10} Minmax & 0.809 \tiny $\pm$ 0.003 & 39.3 \tiny $\pm$ 0.03 & 62.2 \tiny $\pm$ 0.07 \\ \hline
    \end{tabular}}
    \caption{We measure the scalar reward from~\citet{macina-etal-2025-mathtutorbench} as well as the win-rate over sampling and ground-truth responses (GT) on MathDial with TutorRL-7B.
    Bayes provides improvements in all settings.
    Minmax shows degradations due to optimizing for the tutoring strategies \emph{generic} and \emph{telling} too often.
    }
    \label{tab:tab2}
\end{table}

\subsection{Main Results on Tutoring}
Next, we show results on tutor response generation, again using ancestral sampling for $\calH$ without conditioning on a tutoring strategy and using $p(\omega\mid \vx)$ determined by a finetuned Llama-3.2-Instruct model that we train on $90\%$ of the MathDial training data.
The remaining $10\%$ are used for calibration, where the model achieves an accuracy of $47.3\%$ which again shows the high degree of ambiguity.
Since we do not have multiple annotations for the same input we use standard split conformal prediction.
We think ambiguity should be taken into account in future datasets by collecting multiple plausible tutoring strategies and responses for the same input.

Results are shown in~\cref{tab:tab2} and show that, this time, only Bayes decision rule improves results over sampling, though all methods improve over the human ground-truth in terms of reward due to the RL training of TutorRL.
We have analyzed these results in more detail and attribute these to the fact that tutoring strategies that do not always follow pedagogical principles, namely, telling of the solution and generic answers are contained in the prediction sets over $50\%$ of times. As a consequence, minmax seeks to find the highest reward sequence wrt. these strategies, as these minimize utility.
We believe this can be mitigated with stronger strategy predictors but caution that minmax utility seems to depend a lot on the structure of $\Omega$.
    
\subsection{Conformal Prediction Sets}
\label{subsec:results_conformal}
\begin{table}[t]
    \centering
    \resizebox{\linewidth}{!}{\begin{tabular}{l|ccc}
         Generator & Avg. $|\calC| / |\Omega|$ & $\Delta$-Minmax & $\Delta$-Bayes \\ \hline
         & \multicolumn{3}{c}{MathDial} \\\hline
         TutorRL-7B & 0.766 & \textcolor{highlightColor}{-0.049} & \textcolor{linkColor}{+0.034}\\\hline
         & \multicolumn{3}{c}{ICLR} \\\hline
         Gemma-3 & 0.641 & \textcolor{linkColor}{+0.05} & \textcolor{linkColor}{+0.15} \\
         Mistral-3 & 0.641 & \textcolor{linkColor}{+0.21} & \textcolor{linkColor}{+0.34}\\
         Qwen-3 & 0.641 & \textcolor{linkColor}{+0.08}& \textcolor{linkColor}{+0.09}\\\hline
         & \multicolumn{3}{c}{NLPEER} \\\hline
         Gemma-3 & 0.470 & \textcolor{linkColor}{+0.04} & \textcolor{linkColor}{+0.24} \\
         Mistral-3 & 0.470 & \textcolor{linkColor}{+0.14} & \textcolor{linkColor}{+0.35} \\
         Qwen-3 & 0.470 & \textcolor{highlightColor}{-0.13} & \textcolor{linkColor}{+0.05} \\\hline
    \end{tabular}}
    \caption{We compare the compression ratio of conformal sets $\calC$ over the number of possible labels $|\Omega|$ with downstream improvements over sampling in terms of actionability and reward for~\cref{eq:minmax} ($\Delta$-Minmax) and~\cref{eq:bayesrisk_conformal} ($\Delta$-Bayes). Conformal prediction sets are often wide for highly ambiguous tasks which can degrade performance when using minmax strategies.}
    \label{tab:tab3}
\end{table}

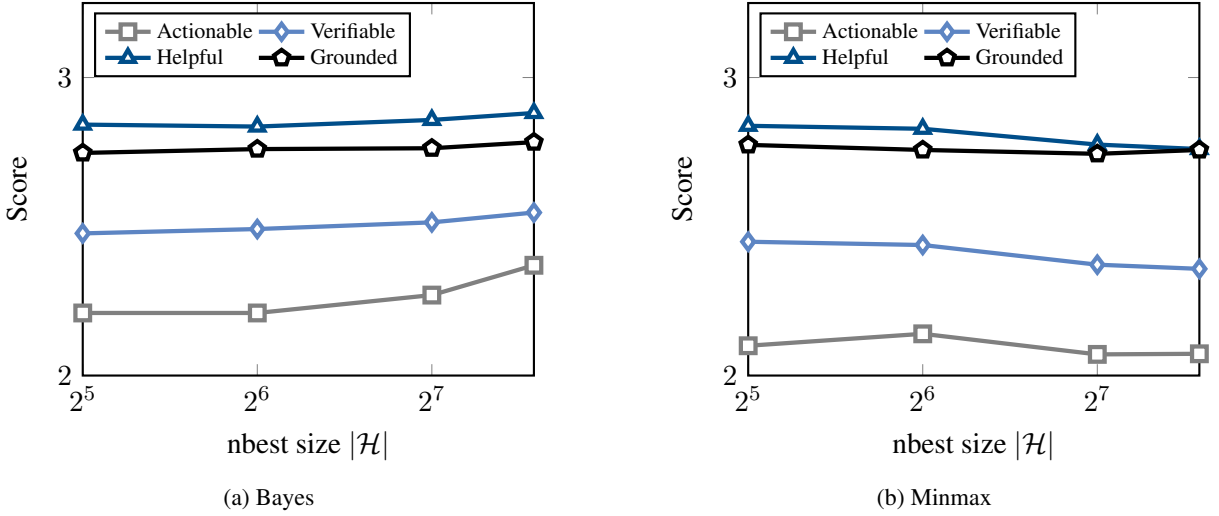
\begin{figure*}[t]
\begin{subfigure}{.45\linewidth}
\centering
\resizebox{\linewidth}{!}{\begin{tikzpicture}
\begin{axis}[
    xlabel=nbest size $|\calH|$,
    ylabel=Score,
    xmin=32.0, xmax=192,
    ymin=2.0, ymax=3.25,
    height=7cm,
    height=6.5cm,
    xtick={32, 64, 128},
    xmode=log,log basis x=2,
    ytick={1.0, 2.0, 3.0, 4.0},
    style=thick,
    xlabel near ticks,
    ylabel near ticks,
    y label style={at={(-0.1,0.5)}},
    legend style={at={(0.025,0.975)}, anchor=north west, nodes={scale=0.75, transform shape}},
    legend cell align={left},
    legend columns=2
            ]
\addplot+[ultra thick,mark=square*,gray, mark size=2.75pt, mark options={scale=1, fill=white}] plot coordinates {
(32, 2.21)
(64, 2.21)
(128, 2.27)
(192, 2.37)
};
\addplot+[ultra thick,mark=diamond*,palette1, mark size=3pt, mark options={scale=1, fill=white}] plot coordinates {
(32, 2.477)
(64, 2.4916)
(128, 2.514)
(192, 2.547)
};
\addplot+[ultra thick,mark=triangle*,palette3, mark size=3pt, mark options={scale=1, fill=white}] plot coordinates {
(32, 2.842)
(64, 2.8355)
(128, 2.8577)
(192, 2.881)
};
\addplot+[ultra thick,mark=pentagon*,black, mark size=3pt, mark options={scale=1, fill=white}] plot coordinates {
(32, 2.747)
(64, 2.7599)
(128, 2.7627)
(192, 2.7832)
};
\addlegendentry{Actionable}
\addlegendentry{Verifiable}
\addlegendentry{Helpful}
\addlegendentry{Grounded}
\end{axis}
\end{tikzpicture}}
\caption{Bayes}
\end{subfigure}\hfill\begin{subfigure}{.45\linewidth}
\centering
\resizebox{\linewidth}{!}{\begin{tikzpicture}
\begin{axis}[
    xlabel=nbest size $|\calH|$,
    ylabel=Score,
    xmin=32.0, xmax=192,
    ymin=2.0, ymax=3.25,
    height=7cm,
    height=6.5cm,
    xtick={32, 64, 128},
    xmode=log,log basis x=2,
    ytick={1.0, 2.0, 3.0, 4.0},
    style=thick,
    xlabel near ticks,
    ylabel near ticks,
    y label style={at={(-0.1,0.5)}},
    legend style={at={(0.025,0.975)}, anchor=north west, nodes={scale=0.75, transform shape}},
    legend cell align={left},
    legend columns=2
            ]
\addplot+[ultra thick,mark=square*,gray, mark size=2.75pt, mark options={scale=1, fill=white}] plot coordinates {
(32, 2.10)
(64, 2.14)
(128, 2.071)
(192, 2.073)
};
\addplot+[ultra thick,mark=diamond*,palette1, mark size=3pt, mark options={scale=1, fill=white}] plot coordinates {
(32, 2.449)
(64, 2.438)
(128, 2.372)
(192, 2.358)
};
\addplot+[ultra thick,mark=triangle*,palette3, mark size=3pt, mark options={scale=1, fill=white}] plot coordinates {
(32, 2.838)
(64, 2.828)
(128, 2.775)
(192, 2.760)
};
\addplot+[ultra thick,mark=pentagon*,black, mark size=3pt, mark options={scale=1, fill=white}] plot coordinates {
(32, 2.774)
(64, 2.757)
(128, 2.744)
(192, 2.757)
};
\addlegendentry{Actionable}
\addlegendentry{Verifiable}
\addlegendentry{Helpful}
\addlegendentry{Grounded}
\end{axis}
\end{tikzpicture}}
\caption{Minmax}
\end{subfigure}
\caption{We provide initial results from scaling the size of $\calH$ using Mistral3 on ICLR and~\cref{eq:bayes_risk} (left) and~\cref{eq:minmax} (right). We find that all criteria are improved for Bayes as $|\calH|$ increases, especially towards the largest size tried ($|\calH| = 192$), but find the opposite trend for minmax which degrades as $\calH$ grows.
}
\label{fig:scaling}
\end{figure*}
Overall, we find in~\cref{tab:tab3} that the prediction sets in a task like tutoring, where uncertainty about the correct strategy is high, are often large.
While these could be narrowed down by allowing a larger error margin, this would also result in less reliable prediction sets.
In reviewing, the prediction sets show a larger compression than in tutoring with cost savings of up to ca. $50\%$, though especially on NLPEER there is a slight performance degradation when using conformal pruning. %
There, results using minmax are better, indicating that the quality of prediction sets is key for the method to work well.

\subsection{Scaling Behavior}
\label{subsec:scaling_behavior}
Finally, we evaluate the scaling behavior of~\cref{eq:bayes_risk} and~\cref{eq:minmax} in terms of $|\calH|$ on Review-5k using Mistral3.
We use sizes ranging from 16 to 192 and find that the Bayes decision rule provides consistent improvements as $\calH$ grows.
The same is not true for the minmax decision rule, where performance even seems to degrade, potentially due to interactions with too wide prediction sets.
Altogether, the experiment provides support for approximations of Bayes decision rule as a method for test-time scaling, which are consistent with prior literature~\citep{daheim2025uncertaintyaware, cong2025improvingloravariationallearning}.

\section{Conclusion \& Future Work}
In this work, we have presented uncertainty-aware methods for decision-making, where decisions consist of an LLM generation, for example, a review or a tutor response, and a categorical decision, for example, a reviewer score or a tutoring strategy.
These methods either use a Bayesian motivation and average over a distribution over the categorical score to maximize expected utility or take a risk-averse approach and maximize utility with respect to a worst-case score within a conformal prediction set, which can also be used to reduce the sum in the Bayesian case to only important examples and can be useful, for example, to understand an expected score distribution or how AI tutors navigate students through complex problems.
We evaluate these strategies in the domains of automatic reviewing and tutoring using various LLMs.
We find that the Bayesian approach always gives improvements while the risk-averse approach can lead to, for example, overly generic responses if the prediction set is wide.
Finally, both can empower practitioners to define their own notions of utility.

Overall, there are many open problems for the community to solve towards making these methods work more reliably, even though we find encouraging first results.
Current LLMs do not always perform well on hard categorical prediction tasks like review score predictions and are prone to providing distributions that do not directly agree with human label variation.
It is also important to account for such label variation when using conformal prediction but this requires actively collecting multiple labelings of the same example, which are not often found in existing datasets.
The methods are also expected to benefit from better LLM judges that, among others, can account for long contexts and judge complex user-defined criteria.
Finally, we believe that there is still work to be done on making LLMs good samplers for such tasks that also reflect plausible diversity of humans.
Our work is therefore best read as a call for action to the community to tackle such hard problems and focus on actively reducing uncertainty during prediction.

\section*{Limitations}
One limitation of our study is that we do not directly know whether any of the data we use for testing was used for training the models used in this study.
The ICLR data is from 2024 and the NLPEER data is from version 1 of the dataset which was released in 2023.
All models that were used are newer than this data and, perhaps, have seen at least some of the reviews during training.
Therefore, we can at least not guarantee that our results are expected to generalize to new papers that are submitted to, say, a conference in 2026.
At the same time, it is necessary to use newer models to ensure that the models perform sufficiently well, otherwise conclusions about a good generation algorithm or decision rule might be driven more by a lack of model capabilities than actual qualities of the algorithms used on top of it.

We do believe that the relevance of our work grows as it is applied to more capable systems than can, perhaps, function as better and more diverse samplers of high-quality reviews.
Unfortunately, though, systems like the one used in NeurIPS are proprietary but we call on the community to also focus on generation algorithms and decision rules.

Finally, our study does not clearly evaluate real-world impact, as there is no field study, for example, on providing feedback before real conference submissions.
Such studies might be interesting but require careful thought.
Similarly, we do not know long-lasting consequences of AI tutor use with a different decision rule, as we could not directly apply our method on real students, which, again, requires careful considerations.

\section*{Ethics Statement}
In this work, we use publicly available research data from the ACL Anthology, OpenReview, and data that was made publicly available through NLPEER via an explicit consent in the ACL Rolling Review.
None of our experiments involve any personal data and only data that has been used extensively in previous research studies.
None of the tutoring experiments involve data from any real students as all students in Mathdial were simulated by LLMs and teachers were crowdworkers who gave explicit consent.
Our work provides methods that can be used to improve automated methods in both peer review and tutoring.
These techniques can bring various benefits to the communities, for example, because they decrease the load of human reviewers, can give authors valuable feedback, and can provide personalized feedback to students.
However, we caution that any adoption of these methods must be carefully prepared and designed.
Before humans are tutored, it needs to be ensured that a tutoring agent follows pedagogical and ethical guidelines.
Before humans receive AI feedback from a reviewing agent, it needs to be ensured that the agent follows sound guidelines and adheres to conference and ethical guidelines.
We emphasize that we do not advocate for using our methods to automate the peer review process but only to use them as support systems that are followed by human supervision.
Similarly, we see them as support to human teachers, for example, for individualized ``any-time'' tutoring but not as a replacement.

\section*{Acknowledgments}
This research work has been funded by the German Federal
Ministry of Research, Technology and Space and the Hessian Ministry of Higher Education, Research, Science and
the Arts within their joint support of the National Research
Center for Applied Cybersecurity ATHENE.

\bibliography{custom}
\newpage
\appendix

\section{Experimental Details}
\label{sec:details}
\subsection{Reviewing Experiments}
We first detail the exact prompts that we use for our work for reproducibility.
The prompts for generating reviews on ICLR and ACL are found in~\cref{prompt:iclr_gen} and~\cref{prompt:acl_gen}, respectively.
These prompts directly follow the guidelines that are posted on OpenReview for human reviewers in order to stay objective.
The models used for generation are exactly detailed in the main body of the text with their corresponding identifier on the huggingface hub (\url{https://huggingface.co/models}).
We use vLLM~\citep{kwon2023efficient} with ancestral sampling and, in particular, do not use any kind of truncation sampling but rather use the full token-level distributions.

The prompts for scoring and utility estimation are given in~\cref{prompt:iclr_score},~\cref{prompt:acl_score},~\cref{prompt:iclr_utility},~\cref{prompt:acl_utility}.
We did not extensively tune the prompts for utility estimation and, in particular, did not optimize them for validation results using the reward model that we use for evaluation.
For both cases, we transform the used models to a classifier in vLLM that outputs a distribution over tokens, for example $0-4$ for 5-point-Likert utility estimation by taking the output embeddings of the corresponding tokens and removing all other output embeddings.
We do this by throwing away all output embeddings but the ones for relevant tokens and then calculating a softmax only over these.
We calculate the final score via an expectation over this classifier.
Denoting it with $p(\cdot\mid \vx)$ and possible scores with $\calS \in \mathbb{N}^{|\calS|}$ we therefore use \begin{equation}
    s(\vy, \omega) = \sum_{s\in\calS} p(s\mid \vx) \cdot s.
\end{equation}
This is consistent with a Bayesian approach where unknowns are marginalized.
The approach has concurrently been dubbed LLM-as-a-verifier~\citep{kwok2026llmverifier} and shown empirical success which we attribute to its inherent uncertainty reduction.

For scoring, we found that zero-shot LLMs without training did not perform well and almost always preferred scores that are between reject and accept or borderline.
We therefore trained a smaller Llama3.2-3B model to predict scores.
On ICLR data, we used the training set that is also provided by~\citet{weng2025cycleresearcher}.
On NLPEER, we use the set of papers with one review in v1 from~\citet{dycke-etal-2023-nlpeer}.
We split off the first $90\%$ of data for training and the remaining $10\%$ for validation.

On ICLR, this amounts to 3,770 training and 419 calibration papers, on NLPEER to 770 training and 86 calibration papers.
The test set on ICLR contains 781 papers and on NLPEER it contains 673 papers.
We train the model for 4 epochs on ICLR and 6 epochs on NLPEER with a batch size of 32 achieved via gradient accumulation.
For both we use AdamW~\citep{loshchilov2018decoupled}, a learning rate of $1e-5$, a small weight decay of $0.000001$, $(\beta_1, \beta_2) = (0.9, 0.99)$, 100 linear warmup steps for the learning rate which then is annealed to 0 using cosine decay.
In both training and prediction we use a maximum input length of $32,768$ tokens for all models.

The evaluation is done by directly using the reward model and scores as defined in~\citet{sahinuc2026reward}.
We use the \textrm{UKPLab/SciRM-Ref-7B} model after splitting the weaknesses section of the paper into individual points using simple rules and apply the model to each weakness and average over them.
A limitation of the approach is that strengths are not included in the rating due to the focus of~\citet{sahinuc2026reward}.
The scaling experiment in~\cref{subsec:scaling_behavior} follows the exact same set-up but varies the size of $\calH$ as described.

All experiments are conducted on NVIDIA A100, RTX Pro 6000 and H200 GPUs with 80GB, 96GB, and 141GB memory.
The experiments take around one day to run for the full pipeline of generation, utility estimation, decision, and evaluation on one of the datasets.
The Llama3.2-3B-Instruct model was trained in ca. 2 hours on one A100 GPU with 80GB memory.

None of the data used contains personally-identifiable information.
We also note that our usage of the data is in accordance with the licenses from~\citet{weng2025cycleresearcher} and~\citet{dycke-etal-2023-nlpeer}.
We only use the data for research purposes and only advocate for usage of the methods within this scope.
We do not promote any use of automated peer review methods in a malicious way or to automate the peer review process, or replace human decision-making.
Rather, we clearly emphasize our goal as helping human reviewers and authors with automated methods to foster a responsible research environment and help all who are included in the peer review process.

\subsection{Tutoring Experiments}
For the tutoring experiments we use TutorRL-7B which is available on the huggingface hub under the identifier \textrm{eth-nlped/TutorRL-7B}.
We do not alter the system prompt but directly use the one defined in the chat template, as this was also used during model training.
We again use ancestral sampling, no truncations of any sort, and use vLLM.

We classify tutoring strategies into \textrm{focus}, \textrm{probing}, \textrm{generic}, and \textrm{telling} using a finetuned Llama3.2-3B model on MathDial using the prompt in~\cref{prompt:mathdial_score}.
We again use AdamW, a learning rate of $1e-5$, a small weight decay of $0.000001$, $(\beta_1, \beta_2) = (0.9, 0.99)$, 100 linear warmup steps for the learning rate which then is annealed to 0 using cosine decay and train for 5 epochs.

For finetuning, we use $90\%$ of the tutor responses from the training set, where a conversation has at least had one utterance, yielding 9,162 utterances.
We calibrate the conformal threshold on the remaining $10\%$ which are 1,018 utterances.
The test set contains 3100 examples.

Utility is calculated with a Qwen3-14B model that again is transformed to a classifier.
We use the prompt in~\cref{prompt:iclr_utility} and again use the LLM-as-a-verifier approach.

For evaluation, we directly use the approach from~\citet{macina-etal-2025-mathtutorbench} using the finetuned model found under \textrm{eth-nlped/Qwen2.5-1.5B-pedagogical-rewardmodel} on the huggingface hub.
We use their input schema and exactly follow the scoring code.

All experiments are conducted on NVIDIA A100, RTX Pro 6000 and H200 GPUs with 80GB, 96GB, and 141GB memory.
The experiments take around two to three hours to run for the full pipeline of generation, utility estimation, decision, and evaluation on one of the datasets.
The Llama3.2-3B-Instruct model was trained in ca. 2 hours on one A100 GPU with 80GB memory.
We average the results over three random seeds.

As MathDial was collected with anonymous crowdworkers and LLM-simulated students, no personally-identifying data is contained or was used by us.

\begin{promptbox}
Given a research paper and the review guidelines below, write a summary of its strengths and weaknesses. Be objective, thoughtful, critical and not too positive. Your points should be grounded in the paper. It is not necessary to balance out the number of strengths and weaknesses. Output a json dictionary.

\#\# Review guidelines

**Summary of Strengths**
What are the major reasons to publish this paper at ICLR? These could include novel and useful methodology, insightful empirical results or theoretical analysis, clear organization of related literature, or any other reason why interested readers of ICLR papers may find the paper useful.

**Summary of Weaknesses**
What are the concerns that you have about the paper that would cause you to favor prioritizing other high-quality papers that are also under consideration for publication? These could include concerns about correctness of the results or argumentation, limited perceived impact of the methods or findings (note that impact can be significant both in broad or in narrow sub-fields), lack of clarity in exposition, or any other reason why interested readers of ICLR papers may gain less from this paper than they would from other papers under consideration. Where possible, please number your concerns as 1., 2., etc. so authors may respond to them individually.

\#\# Output format
Output only the json dictionary and follow the json schema exactly, with no extra keys, notes, comments, or explanations:
{"strengths": "...", "weaknesses": "..."}
\end{promptbox}
\captionof{figure}{Prompt used to generate reviews for ICLR papers.}\label{prompt:iclr_gen}

\begin{promptbox}
Given a research paper and the review guidelines below, write a summary of its strengths and weaknesses. Be objective, thoughtful, critical and not too positive. Your points should be grounded in the paper. It is not necessary to balance out the number of strengths and weaknesses. Output a json dictionary.

\#\# Review guidelines

**Summary of Strengths**
What are the major reasons to publish this paper at a selective *ACL venue? These could include novel and useful methodology, insightful empirical results or theoretical analysis, clear organization of related literature, or any other reason why interested readers of *ACL papers may find the paper useful.

**Summary of Weaknesses**
What are the concerns that you have about the paper that would cause you to favor prioritizing other high-quality papers that are also under consideration for publication? These could include concerns about correctness of the results or argumentation, limited perceived impact of the methods or findings (note that impact can be significant both in broad or in narrow sub-fields), lack of clarity in exposition, or any other reason why interested readers of *ACL papers may gain less from this paper than they would from other papers under consideration. Where possible, please number your concerns as 1., 2., etc. so authors may respond to them individually.

\#\# Output format
Output only the json dictionary and follow the json schema exactly, with no extra keys, notes, comments, or explanations:
{"strengths": "...", "weaknesses": "..."}
\end{promptbox}
\captionof{figure}{Prompt used to generate reviews for ACL papers.}\label{prompt:acl_gen}

\begin{promptbox}
Given a review and scoring guidelines below, return a single number from the guidelines to indicate a score for a research paper that is consistent with the review.
Be objective. A large number of strengths and few weaknesses indicate a good score. A large number of weaknesses and few strengths indicate a bad score. A similar number of both might be a borderline paper.
\#\# Possible Scores

9: Top-quality paper: Top 1\%
7-8: Top-quality paper: Top 5\%
5-6: Accept: Strong paper with good contribution.
4: Weak Accept: Borderline paper, likely to be accepted.
3: Marginally below the acceptance threshold: Would not mind if the paper is accepted.
2: Weak Reject: Borderline paper, likely to be rejected.
0-1: Reject: Poor or deeply flawed paper

\#\# Output format
Output only one number
\end{promptbox}
\captionof{figure}{Prompt used to score ICLR papers.}\label{prompt:iclr_score}

\begin{promptbox}
Given a review and scoring guidelines below, return a single number from the guidelines to indicate a score for a research paper that is consistent with the review.
Be objective. A large number of strengths and few weaknesses indicate a good score. A large number of weaknesses and few strengths indicate a bad score. A similar number of both might be a borderline paper.
\#\# Possible Scores

9 = Top-Notch: This is one of the best papers I read recently, of great interest for the (broad or narrow) sub-communities that might build on it
8
7 = This paper represents solid work, and is of significant interest for the (broad or narrow) sub-communities that might build on it
6
5 = Good: This paper makes a reasonable contribution, and might be of interest for some (broad or narrow) sub-communities, possibly with minor revisions
4
3 = Revisions Needed: This paper has some merit, but also significant flaws, and needs work before it would be of interest to the community
2
1 = Major Revisions Needed: This paper has significant flaws, and needs substantial work before it would be of interest to the community
0 = This paper is not relevant to the *ACL community (for example, is in no way related to natural language processing)

\#\# Output format
Output only one number
\end{promptbox}
\captionof{figure}{Prompt used to score ACL papers. Note that we have transformed the scores to map to single number tokens for simplicity.}\label{prompt:acl_score}

\begin{promptbox}
Given are a review and two guidelines.
The first guideline explains what score should be given to a paper.
The second guideline explains how to score the review for the paper based on review and score.

Your task is to score the review.

\#\# Paper Score Guideline

9: Top-quality paper: Top 1\%
7-8: Top-quality paper: Top 5\%
5-6: Accept: Strong paper with good contribution. Typically many strengths and some weaknesses.
4: Weak Accept: Borderline paper, likely to be accepted. Typically similar amount of strengths and weaknesses but strengths outweigh weaknesses.
3: Marginally below the acceptance threshold: Would not mind if the paper is accepted. Typically similar amount of strengths and weaknesses but strengths outweigh weaknesses slightly.
2: Weak Reject: Borderline paper, likely to be rejected. Typically more weaknesses than strengths.
0-1: Reject: Poor or deeply flawed paper. Typically a paper with many weaknesses that can not easily be resolved.

\#\# Review Score Guideline

0: The review is not consistent with the paper score or of low quality. For example, the review could be very positive and the score low, or the review could be negative but the score high. A low quality review is indicated by unsubstantiated claims, feedback that is not actionable, helpful, or verifiable.
1: The review is only somewhat consistent with the review score or of rather low quality. For example, the review contains many unsubstantiated claims and is unfair, contains only little actionable feedback, is not very helpful, and many points are hard to verify.
2: The review fits the score to some extent and is of average quality. For example, it is not in-depth or contains some unsubstantiated claims, some claims that are not actionable, is somewhat helpful, and has a few points that are hard to verify. Some points are grounded in the paper.
3: The review and score fit. The review has at most very few unsubstantiated claims, fairly addresses the papers weaknesses, and provides actionable guidelines to the authors. The review is helpful and most points are grounded in the paper.
4: The review and score fit. The review has no unsubstantiated claims, fairly addresses the papers weaknesses, and provides actionable guidelines to the authors. The review is of exceptional quality. All points are grounded well in the paper. 

\#\# Output Format
Return only one score in the range 0-4 from the review score guidelines and nothing else.
\end{promptbox}
\captionof{figure}{Prompt for estimating the utility of ICLR reviews used in $u(\vy, \omega)$.}\label{prompt:iclr_utility}

\begin{promptbox}
Given are a review and two guidelines.
The first guideline explains what score should be given to a paper.
The second guideline explains how to score the review based on review and score.

Your task is to score the review objectively.

\#\# Paper Score Guideline

9 = Top-Notch: This is one of the best papers I read recently, of great interest for the (broad or narrow) sub-communities that might build on it
8
7 = This paper represents solid work, and is of significant interest for the (broad or narrow) sub-communities that might build on it
6
5 = Good: This paper makes a reasonable contribution, and might be of interest for some (broad or narrow) sub-communities, possibly with minor revisions
4
3 = Revisions Needed: This paper has some merit, but also significant flaws, and needs work before it would be of interest to the community
2
1 = Major Revisions Needed: This paper has significant flaws, and needs substantial work before it would be of interest to the community
0 = This paper is not relevant to the *ACL community (for example, is in no way related to natural language processing)

\#\# Review Score Guideline

0: The review is not consistent with the paper score or of low quality. For example, the review could be very positive and the score low, or the review could be negative but the score high. A low quality review is indicated by unsubstantiated claims, feedback that is not actionable, helpful, or verifiable.
1: The review is only somewhat consistent with the review score or of rather low quality. For example, the review contains many unsubstantiated claims and is unfair, contains only little actionable feedback, is not very helpful, and many points are hard to verify.
2: The review fits the score to some extent and is of average quality. For example, it is not in-depth or contains some unsubstantiated claims, some claims that are not actionable, is somewhat helpful, and has a few points that are hard to verify. Some points are grounded in the paper.
3: The review and score fit. The review has at most very few unsubstantiated claims, fairly addresses the papers weaknesses, and provides actionable guidelines to the authors. The review is helpful and most points are grounded in the paper.
4: The review and score fit. The review has no unsubstantiated claims, fairly addresses the papers weaknesses, and provides actionable guidelines to the authors. The review is of exceptional quality. All points are grounded well in the paper. 

\#\# Output Format
Return only one score in the range 0-4 from the review score guidelines and nothing else.
\end{promptbox}
\captionof{figure}{Prompt for estimating the utility of ACL reviews used in $u(\vy, \omega)$.}\label{prompt:acl_utility}

\begin{promptbox}
    Explicit actions or suggestions are direct or apparent. Authors can directly identify modifications they should apply to their draft. Clarification questions should be treated as explicit statements if they give a direct action. However, implicit actions need to be inferred from the comment. This includes missing parts that need to be added. Authors can deduce what needs to be done after reading the comment. For concrete actions, the authors know exactly what needs to be done and how to apply the action. However, for vague actions the authors still don’t know how to carry out this action. Scoring rubric is as follows:\\1: The comment lacks meaningful information to help authors improve the paper. Authors do not know what they should do after reading the comment.\\3: The comment explicitly states an action but is vague on how to execute it.\\5: The comment contains an explicit action and concrete details on how to implement it. Authors know exactly how to apply it.
\end{promptbox}
\captionof{figure}{Criteria used to evaluate the actionability of generated reviews on a Likert scale. The criteria are taken from~\citet{sahinuc2026reward} and used together with the system prompt and five examples from their work, which are left out for brevity.}\label{prompt:review_eval_actionability}

\begin{promptbox}
    Claim justification-verification can be done either by logical reasoning supporting the claim, common sense knowledge in the field verifying the claim (e.g., referencing established practices or standards), or external references substantiating the claim. Scoring rubric is as follows:\\1: The comment contains a claim without any supporting evidence or justification.\\2: The comment provides some support for its claim, but the justification is vague, insufficient, or not fully articulated. Authors may struggle to follow the reasoning.3: The comment provides support for its claim, but key elements are missing, such as specific examples, detailed explanations, or supporting references. Authors must make a significant effort to follow the justification.\\4: The comment’s claim is sufficiently supported but has minor gaps. The reviewer could provide a more detailed explanation or reference.\\5: The claim is thoroughly supported by explicit, sufficient, and robust evidence. This can be achieved through: - Clear and precise reasoning or explanation. - Specific and relevant references to external works or data. - Logical and unassailable common-sense arguments.
\end{promptbox}
\captionof{figure}{Criteria used to evaluate the verifiability of generated reviews on a Likert scale. The criteria are taken from~\citet{sahinuc2026reward} and used together with the system prompt and five examples from their work, which are left out for brevity.}\label{prompt:review_eval_verifiability}

\begin{promptbox}
    For fully grounded comment, the author can accurately pinpoint the section, table, figure, or unique aspect being addressed. For weak grounded comment, the author can make an educated guess but cannot precisely identify the referenced part. For specificity, the comment should detail what is wrong or missing in the referenced part. If external work is mentioned, it should also provide specific examples. Scoring rubric is as follows:\\1: The comment is not grounded at all. It does not identify a specific area in the paper. The comment is highly unspecific.\\2: The authors cannot confidently determine which part the comment addresses. Further, the comment does not specify what needs to be addressed in this part.\\3: The authors cannot confidently determine which part the comment addresses. However, the comment clearly specifies what needs to be addressed in this part.\\4: The comment explicitly mentions which part of the paper it addresses, or it should be obvious to the authors. However, this comment does not specify what needs to be addressed in this part.\\5: The comment explicitly mentions which part of the paper it addresses, and it is obvious to the authors. The comment specifies what needs to be addressed in this part.
\end{promptbox}
\captionof{figure}{Criteria used to evaluate the groundedness of generated reviews on a Likert scale. The criteria are taken from~\citet{sahinuc2026reward} and used together with the system prompt and five examples from their work, which are left out for brevity.}\label{prompt:review_eval_grounded}

\begin{promptbox}
    A helpful review should be actionable, grounded on a specific part of the paper, provide justification or evidence to its claims. Scoring rubric is as follows:\\1: The comment fails to identify meaningful weaknesses or suggest improvements, leaving the authors with no actionable feedback.\\2: The comment identifies a weakness or improvement area but is vague, lacks clarity, or provides minimal guidance, making it only slightly beneficial for the authors.\\3: The comment identifies weaknesses or areas for improvement but is incomplete or lacks depth. While the authors gain some insights, the feedback does not fully address their needs for improving the draft.\\4: The comment provides clear and actionable feedback on weaknesses and areas for improvement, though it could be expanded or refined to be fully comprehensive and impactful.\\5: The comment thoroughly identifies weaknesses and offers detailed, actionable, and constructive suggestions that empower the authors to significantly improve their draft.
\end{promptbox}
\captionof{figure}{Criteria used to evaluate the helpfulness of generated reviews on a Likert scale. The criteria are taken from~\citet{sahinuc2026reward} and used together with the system prompt and five examples from their work, which are left out for brevity.}\label{prompt:review_eval_helpful}

\begin{promptbox}
You are given a conversation between a student and a teacher, where the teacher attempts to guide the student towards correcting their incorrect solution attempt.
Your task is to indicate the tutoring strategy that the teacher should use in their next utterance from the following choices.
DO NOT CLASSIFY THE CURRENT TEACHER UTTERANCE.
Follow the guidelines below, where there is an explanation of intents behind each strategy with examples.

0. Focus:
    - Seek Strategy: So what should you do next?
    - Guiding Student: Can you calculate . . . ?
    - Recall Relevant Information: Can you reread the question and tell me what is . . . ?

1. Probing:
    - Asking for Explanation: Why do you think you need to add these numbers?
    - Seeking Self Correction: Are you sure you need to add here?
    - Perturbing the Question: How would things change if they had . . . items instead?
    - Seeking World Knowledge: How do you calculate the perimeter of a square?

2. Telling:
    - Revealing Strategy: You need to add . . . to . . . to get your answer.
    - Revealing Answer No, he had . . . items.

3. Generic:
    - Greeting/Farewell: Hi . . . , how are you doing with the word problem?
    - General inquiry: Can you go walk me through your solution?

Telling should only be used rarely, for example, if the student is stuck.

\#\# Output Format
Return only one number in the range 0-3 to indicate the tutoring strategy from the list and nothing else.
\end{promptbox}
\captionof{figure}{Prompt for classifying tutor strategies.}\label{prompt:mathdial_score}

\begin{promptbox}
You are given a conversation between a student and a teacher, where the teacher attempts to guide the student towards correcting their incorrect solution attempt, a guideline for tutoring strategies and a guideline for scoring.
Your task is to predict the quality of the last reponse generated by the teacher model.

\# Tutoring Strategies

0. focus:
    - Seek Strategy: So what should you do next?
    - Guiding Student: Can you calculate . . . ?
    - Recall Relevant Information: Can you reread the question and tell me what is . . . ?

1. probing:
    - Asking for Explanation: Why do you think you need to add these numbers?
    - Seeking Self Correction: Are you sure you need to add here?
    - Perturbing the Question: How would things change if they had . . . items instead?
    - Seeking World Knowledge: How do you calculate the perimeter of a square?

2. telling:
    - Revealing Strategy: You need to add . . . to . . . to get your answer.
    - Revealing Answer No, he had . . . items.

3. generic:
    - Greeting/Farewell: Hi . . . , how are you doing with the word problem?
    - General inquiry: Can you go walk me through your solution?

\# Scoring Guidelines

Use these criteria to score a response according to the following rubrics:
(a) correctness: the teacher should guide the student towards the correct answer and not state incorrect facts
(b) scaffolding instead of giving away the answer: the teacher should help the student to cognitively engage with the problem and discover the answer on their own
(c) encourage self-correction: by correctly identifying the student mistake and first giving the student the opportunity to self-correct and learn from a mistake 
(d) not overloading the student: manage cognitive load by not giving too much information at once.

0: The response is either inappropriate, does not fit to the given tutoring strategy or does not satisfy any of the above criteria.

1: The response tells the solution too early, or does not provide the student with any actionable feedback or room to improve their understanding. Overall, it only satisfies few of the criteria.

2: The response has good pedagogical quality and provides the student with some room to improve their understanding. It satisfies most of the criteria but not all of them. It fits the tutoring strategy but not perfectly.

3: The response is of high pedagogical quality. It satisfies almost all criteria above and fits the tutoring strategy but is not completely tailored to it. It does not tell the solution directly, unless at the end of the conversation.

4: The response is of exceptional pedagogical quality. It does not tell the solution directly unless absolutely necessary. It provides the student with explicit ways to improve their understanding. It satisfies all of the above criteria. The response is tailored to the tutoring strategy.

\#\# Output Format
Return only one number in the range 0-4 to indicate the quality of the teacher response according to the Scoring Guidelines and nothing else.
\end{promptbox}
\captionof{figure}{Prompt for estimating the utility of tutor responses used in $u(\vy, \omega)$. The examples are taken verbatim from~\citet{macina-etal-2023-mathdial}.}\label{prompt:mathdial_utility}

\begin{promptbox}
    Judge the pedagogical quality of the responses provided by two teachers. Focus on the quality of the scaffolding guidance, correctness, and actionability of the feedback through nudges, questions and hints. Do not give high scores for revealing the full answer.
\end{promptbox}
\captionof{figure}{Prompt used for evaluating tutor respones, taken from~\citet{macina-etal-2025-mathtutorbench}.}\label{prompt:mathdial_eval}

\end{document}